\documentclass[conference]{IEEEtran}
\IEEEoverridecommandlockouts
\usepackage{cite}
\usepackage{amsmath,amssymb,amsfonts}
\usepackage{algorithmic}
\usepackage{graphicx}
\usepackage{textcomp}
\usepackage{xcolor}
\def\BibTeX{{\rm B\kern-.05em{\sc i\kern-.025em b}\kern-.08em
    T\kern-.1667em\lower.7ex\hbox{E}\kern-.125emX}}
    
\newcommand{\RomanNumeralCaps}[1]
    {\MakeUppercase{\romannumeral #1}}    

\begin{document}

\title{Unsupervised Synthesis of Anomalies in Videos:\\ Transforming the Normal\\
\thanks{\textbf{Project webpage:} \textit{https://abhjoshi8.github.io/VideoAnomalySynthesis}}
}

\author{\IEEEauthorblockN{Abhishek Joshi}
\IEEEauthorblockA{IIT Kanpur \\
Dept. of Computer Science\\
\tt\small abhjoshi@iitk.ac.in}
\and
\IEEEauthorblockN{Vinay P. Namboodiri}
\IEEEauthorblockA{IIT Kanpur \\
Dept. of Computer Science\\
\tt\small vinaypn@iitk.ac.in}
}

\maketitle

\begin{abstract}
Abnormal activity recognition requires detection of occurrence of anomalous events that suffer from a severe imbalance in data. In a video, normal is used to describe activities that conform to usual events while the irregular events which do not conform to the normal are referred to as abnormal. It is far more common to observe normal data than to obtain abnormal data in visual surveillance. In this paper, we propose an approach where we can obtain abnormal data by transforming normal data. This is a challenging task that is solved through a multi-stage pipeline approach. We utilize a number of techniques from unsupervised segmentation in order to synthesize new samples of data that are transformed from an existing set of normal examples. Further, this synthesis approach has useful applications as a data augmentation technique. An incrementally trained Bayesian convolutional neural network (CNN) is used to carefully select the set of abnormal samples that can be added. Finally through this synthesis approach we obtain a comparable set of abnormal samples that can be used for training the CNN for the classification of normal vs abnormal samples. We show that this method generalizes to multiple settings by evaluating it on two real world datasets and achieves improved performance over other probabilistic techniques that have been used in the past for this task.
\end{abstract}
\textbf{\textbf{}}

\section{Introduction}
Visual surveillance generates a huge amount of data. A single camera alone can generate terabytes of data over time. This data needs to be screened for abnormal events. This is usually done either through operators that keep active lookout on various screens or as a means of deterrence where the videos are analyzed after the effect if an untoward incidence (such as robbery) has occurred. Clearly, there is a need to automate this tedious process. Therefore, identifying abnormal activities automatically has been an important problem from the application perspective. The task however is challenging as abnormal activity is a highly class imbalance problem in a supervised setting. A number of problems have been solved by the community using deep learning techniques that have been trained using supervision. However, to adopt this approach a large set of labels need to be explicitly provided for abnormal activity.

This is a challenge as usually the surveillance cameras typically capture the normal activity and abnormal occurrences are rare. A large number of these rare events would need to be explicitly annotated. Sometimes, an activity is unusual because there are no past occurrences and evidences of that activity. So, the data describing abnormal activities is scarce and we do not have much labeled data available for them. Hence, it would be desirable to have a video surveillance algorithm which leverages only data of normal activity to make a decision as to whether a particular activity is normal or abnormal. 
\begin{figure}[t]
\begin{minipage}[t]{0.49\linewidth}
    \includegraphics[width=\linewidth]{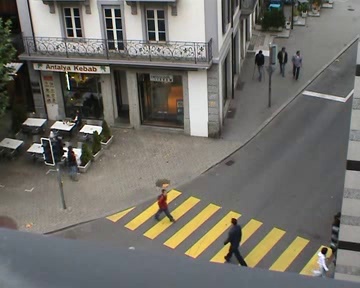}
    \centering
    \label{f1}
\end{minipage}%
    \hfill%
\begin{minipage}[t]{0.49\linewidth}
    \includegraphics[width=\linewidth]{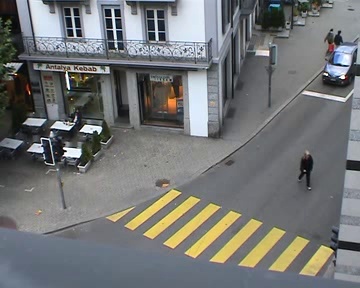}
  
    \label{f2}
\end{minipage} 

\small{(a) Traffic Junction Dataset \cite{varadarajan2009topic}. The left image is normal while the right is an anomaly with a person crossing the road away from the zebra crossing} 



\begin{minipage}[t]{0.49\linewidth}
    \includegraphics[width=\linewidth]{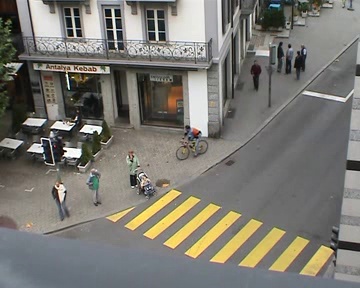}
    \centering
    \label{f3}
\end{minipage}%
    \hfill%
\begin{minipage}[t]{0.49\linewidth}
    \includegraphics[width=\linewidth]{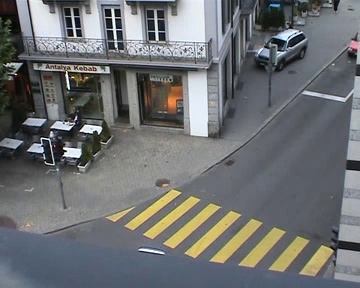}
    \label{f4}
\end{minipage} 

\small{(b) Traffic Junction Dataset \cite{varadarajan2009topic}. The left image is normal while the right is an anomaly as the car enters the pedestrian area.} 

\begin{minipage}[t]{0.49\linewidth}
    \includegraphics[width=\linewidth]{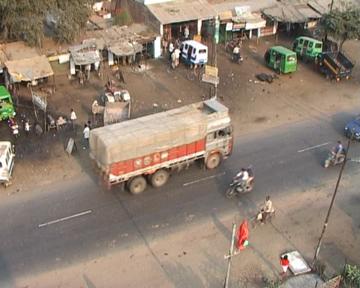}
    \centering
    \label{f5}
\end{minipage}%
    \hfill%
\begin{minipage}[t]{0.49\linewidth}
    \includegraphics[width=\linewidth]{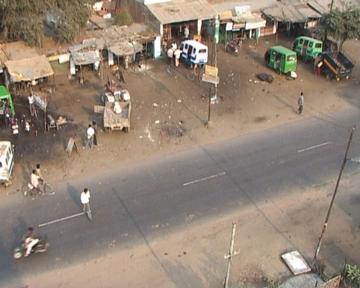}
    \label{f6}
\end{minipage} 

\small{(c) Highway Dataset \cite{pathak2015anomaly}. The left image is normal while the right is an anomaly as a jaywalker crosses the road.}

\caption{Sample Normal and Abnormal Frames from surveillance video datasets.}
\label{traffic-dataset}
\vspace{-4mm}
\end{figure}
In this work, we follow this approach and transform the normal observations to automatically obtain abnormal observations.

In this paper, we follow a synthesis based approach to obtain the supervision. There have been instances of interesting work that have used this approach based on synthesis. For instance, Rozantsev {\it et al.} \cite{fua_synthetic} pursue the generation of synthetic data based on 3D models that are used for training object detectors. Very recent work pursued by Shrivastava {\it et al.} \cite{apple_synthetic} obtains increased realism in generation of synthetic data using generative adversarial networks (GANs). However, the previous approaches obtained synthetic examples by modeling the 3D model and generating realistic renderings of the same. The challenge in our case is that we have samples of normal data but choose to not use any examples of abnormal data (as a few chosen abnormal examples would be limiting and not span the class of abnormal examples). In our approach, we adopt a different architecture where we use a synthesis technique that can generate large number of modified samples from normal samples to obtain abnormal samples. We then use a trained Bayesian network (compared also to non-Bayesian network) that assumes that the synthesized examples are fake and uses them to {\it choose} a set of examples that are abnormal and can be used. 

In Section \RomanNumeralCaps{2} we discuss a number of approaches adopted in the literature. In Section \RomanNumeralCaps{3} we describe the methodology followed by us i.e. the unsupervised setting for synthesizing anomaly and abnormal activity detection. In Section \RomanNumeralCaps{4} we provide experiments and compare our results with the unsupervised models that have been previously proposed in \cite{varadarajan2009topic} and~\cite{pathak2015anomaly}. We provide analysis in Section \RomanNumeralCaps{5} and finally conclude with a discussion in Section \RomanNumeralCaps{6}.

\section{Related Work}

There are broadly two main approaches that have been followed to solve abnormal activity detection in videos. The first approach involves tracking objects in the video frames. The deviation in trajectory points leads to potential candidates for being abnormal \cite{hu2007semantic} \cite{le2008query} \cite{zhang2013mining}. Though these methods perform well, they are subject to trajectory abnormality constraints that may not be prevalent. For instance, due to occlusions there may be a number of truncated or abnormal trajectories obtained from normal sequences. Further, some abnormal sequences, due to short trajectories may appear normal. Moreover, a number of times, it is not the trajectory itself that is abnormal but the location where it occurs that makes it abnormal. For instance as illustrated in Figure \ref{traffic-dataset}(a), the act of crossing the road may appear normal but is abnormal, if not done on the pedestrian crossing. 
The second approach leverages feature descriptors to procure intrinsic patterns of the events, which are eventually used to model the behaviour. Although, absence of tracking information in the latter approach may lead to substantial loss but this approach shows a more convincing way to develop more generic and real-life models. We follow this broad category of approach for our task. Niebles \textit{et al.}~\cite{niebles2008unsupervised} apply topic models for the task of action recognition and classification by modeling the features in terms of visual words. In an unsupervised way, they use pLSA-LDA models to predict the actions.
  Sparse representation has been used in \cite{lu2013abnormal} \cite{zhao2011online} to learn the dictionary of normal behaviours. The behaviours which have large reconstruction errors are considered as anomalous behaviors during testing. Recently, a weakly supervised setting has been proposed in \cite{Sultani_2018_CVPR} which considers normal and anomalous videos as bags and video segments as instances in multiple instance learning (MIL) and predicts anomalous videos.
 
Deep Learning approaches have lead to successes in many computer vision tasks \cite{krizhevsky2012imagenet} \cite{girshick2015fast} including anomaly detection. Xu \textit{et al.} \cite{xu2015learning} demonstrate the effectiveness of deep learning features through a multi-layer auto-encoder for feature learning. In the work \cite{hasan2016learning}, Hasan \textit{et al.} propose a 3D convolutional auto-encoder to model normal frames, which have been further boosted in \cite{chong2017abnormal} \cite{luo2017remembering} via both appearance and motion based model. Recently, Luo \textit{et al.} \cite{luo2017revisit} propose a temporally coherent sparse coding based method which maps to a stacked RNN framework. It is interesting to note that all these anomaly detection approaches are based on the reconstruction of normal training data with the assumption that abnormal events would correspond to larger reconstruction errors.
  
Varadarajan \textit{et al.} \cite{varadarajan2009topic} adopt topic modelling for scene analysis and detection of abnormal events occurring in the video. The assumption is that in a domain, the set of usual events is fixed and can be extracted from the distribution of the visual words and the video clips in the domain. The idea is to build a model for the normal activities in the videos. A video clip having instance of abnormal events would then be expected to have a low likelihood over the model learnt from the normal events. Pathak \textit{et al.} \cite{pathak2015anomaly} propose a mechanism to extend the topic-based analysis of anomalous documents and is combined with a classifier based on spatio-temporal quantized words. As these are the main related approaches for our method, we benchmark our approach by comparing against these. Further, our approach uses a synthesis based transformation approach for generating abnormal samples that has to the best of our knowledge not been adopted for this problem.

\subsection{Background}

\begin{figure*}[t]
\includegraphics[ width=0.9\linewidth]{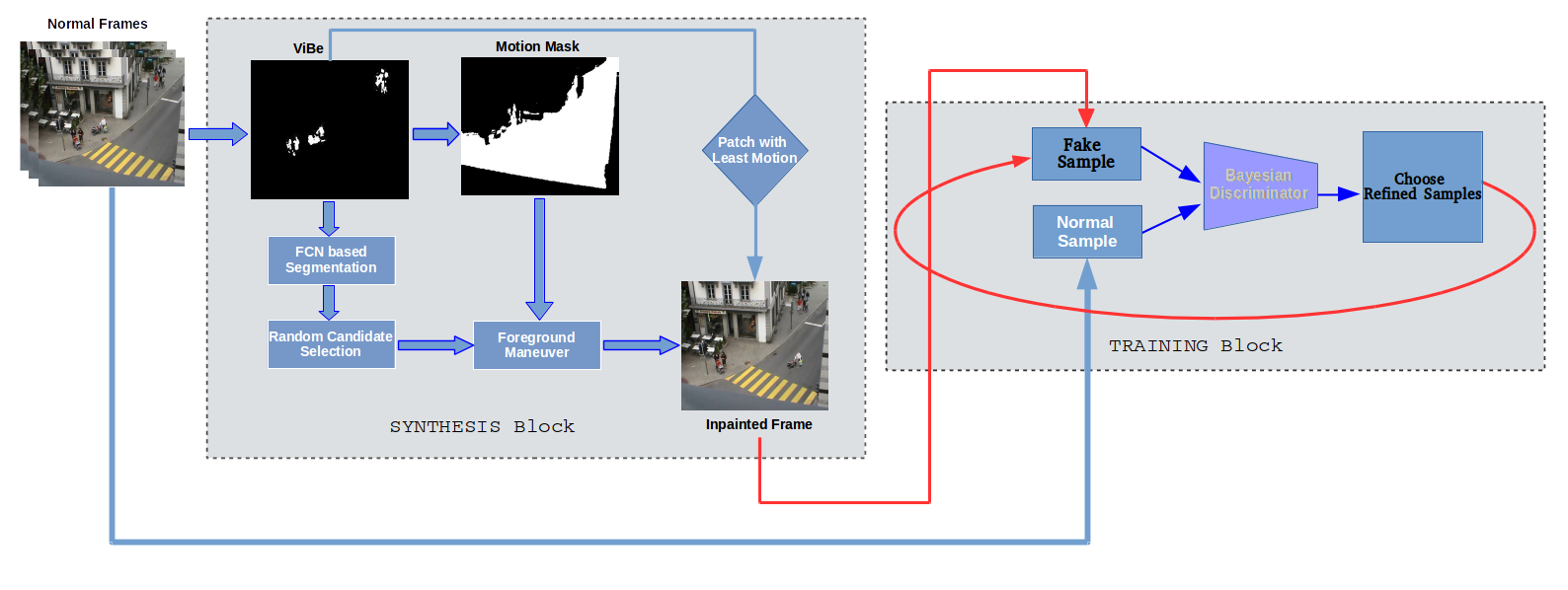}
    \centering
    \caption{Illustrative overview of the proposed approach. There are two main blocks, a Synthesis Block and a Training Block.}
    \label{fig:block-diagram}
\end{figure*}

A crucial component in our model is the use of a Bayesian CNN for obtaining a discriminator. For the sake of completeness we present the related background for obtaining a Bayesian CNN. In probabilistic modeling, we infer a distribution over parameters $w$ of a function $y^*=f(x^*;w)$, that is likely to generate our outputs. Also it is assumed that the training observations, input $X=\{ {x_1,x_2,...x_N}\} $ and their corresponding outputs $Y=\{ {y_1,y_2.....y_N}\}$ are random variables from a distribution. The function $f$ represents the network architecture, and $w$ is the collection of the model parameters. 
Our objective is to model the uncertainty in prediction for classifying our synthesized abnormal samples.
One of the popular ways to do this is the Bayesian approach. Given observation $X,Y$, we need to find the posterior distribution over space of functions, i.e., $p(w|X,Y)$, which captures the most likely function parameters, given our observed data.  The posterior can be modeled with the help some prior distribution $p_0(w)$  over the space of parameters $w$ and with the likelihood function $p(Y|X,w)$. The likelihood function is the probability of the synthesized sample $Y$ given the function parameter $w$. Then, we can predict the output of a new data point $x^*$ by integrating over all possible function parameters $w$. So the predictive distribution~\cite{Gal_ICML2016} \cite{Li_ArXiv2017} is given by the following equation.
\vspace{-1mm}
\begin{equation}
     p(y^*|x^*,X,Y)= \int{p(y^*|x^*,w)p(w|X,Y)dw}
     \label{eq1}
\end{equation}

\vspace{-1mm}

The posterior distribution $p(w|X,Y)$ in eqn.~\ref{eq1} is intractable. To approximate the intractable posterior distribution $p(w|X,Y)$, we need to define an approximating variational distribution $q_{\theta}(w)$ parameterized by $\theta$, whose structure is easy to evaluate. We thus minimize the Kullback–Leibler(KL) divergence~\cite{Kullback_AMS1951} between approximate posterior $q_{\theta}(w)$ and the full posterior $p(w|X,Y)$ w.r.t $\theta$, which is denoted by $ KL(q_{\theta}(w) || p(w|X, Y))$. Minimizing the KL divergence is equivalent to maximizing the log evidence lower bound~\cite{Bishop_Springer2006} with respect to the variational parameters defining $q_{\theta}(w)$,
\begin{equation}
 \label{eq2}
Loss =  \int{q_{\theta}(w)} \log {p(Y|X,w)}dw - KL(q_{\theta}(w)||p_{0}(w))\\
\end{equation}

The posterior $p(w|X,Y)$ in equation ~\ref{eq1} is replaced with approximate posterior $q_{\theta}(w)$. The integral of predictive distribution~\cite{Gal_ICML2016} is intractable for many models because it is integrated over all possible values of $w$. To approximate it, we could condition the model on a finite set of parameters $w$. So we approximate the integral with Monte Carlo integration.  
\begin{equation}
\label{eq6}
\begin{split}
    p(y^*=c|x^*,X,Y) &= \int{p(y^*=c|x^*,w)p(w|X,Y)dw}\\ 
                &\approx \int{p(y^*=c|x^*,w)q_{\theta}(w)dw}\\ 
                &\approx \frac{1}{M}\sum_{m=1}^{M}{p(y^*=c|x^*,\Hat{w}_{m})}
    \end{split}
\end{equation}
\vspace{-3mm}
\noindent with $\Hat{w}_{m}\sim q_{\theta}(w)$, which is the dropout distribution.

%



\section{Methodology}

Our aim is to synthesize frames in an unsupervised way, which appear to be abnormal leveraging solely the normal frames, to achieve detection of abnormal activities in a video. We discuss the intuition and techniques for the desired task in further subsections. As illustrated in Figure \ref{fig:block-diagram}, our overall pipeline mainly consists of two components: \textit{Synthesis Block} and \textit{Training Block}. 

Given the normal videos frames as input, the task of \textit{Synthesis Block} is to synthesize abnormal or \textit{fake} frames by transforming the normal. This is achieved in several stages as shown in Figure \ref{fig:block-diagram}. Firstly, through ViBe \cite{ViBe} technique we obtain foreground masks of moving objects. These noisy segments act as \textit{pseudo ground truths} for training a fully convolutional network (FCN) \cite{FCN} to obtain better foreground through semantic segmentation. Simultaneously, we keep track of the motion of foreground objects to obtain a representative map for overall motion in the video. Next, we randomly sample and cut a candidate foreground object and with the representative motion mask's assistance, the foreground object is maneuvered to a different place in the frame. In order to obtain a neat and indistinguishable fake frame, the original patch is algorithmically replaced with an appropriate patch from the prior frames. Subsection \RomanNumeralCaps{3}-\textit{A} explains the entire mechanism of the \textit{Synthesis Block} in detail. 

The role of the \textit{Training Block} is concisely that of a discriminator. Once fake frames have been synthesized, the challenge further lies in determining which ones among those are more likely to be considered \textit{abnormal}. This block is responsible for selecting indistinguishable \textit{fake} samples generated from the \textit{Synthesis Block}, through incremental training to build a robust classifier. To serve the purpose, our idea is to incrementally train a Bayesian CNN, model the uncertainty and use it to classify abnormal frames. Based on the predictive posterior probabilities obtained from the classifier, we sample more \textit{fake} abnormal ones and iteratively keep adding them to the training set. The methodology is further described in Section \RomanNumeralCaps{3}-\textit{B} 

\subsection{Unsupervised Setting: Synthesizing Anomaly}
In order to obtain realistic anomalous frames we rely on the following principles, a) abnormal instances would be based on unusual location of entities (car on a pedestrian pathway, pedestrian in the middle of the road). b) realistic synthesis would require precise extraction through segmentation and realistic replacement of the extracted segment. 

We rely on motion cues for unsupervised segmentation of moving entities and further rely on this cue to obtain a motion map to evaluate possible placement and segment replacement cues. These are obtained without supervision. This is not unusual as  extensive human vision studies show that motion plays an important role in the development of human visual system. For instance, it has been shown experimentally \cite{Blindness} that soon after gaining sight humans are better able to categorize objects that are seen to be in motion than those seen to be at rest. A similar observation has been studied for infants as well as described in \cite{Infants}. 

\subsubsection{Unsupervised Motion Segmentation}
In a real world scenario, almost all anomalous events occur due to objects in motion. For example, in a traffic road scene as shown in Figure \ref{traffic-dataset}, the abnormal events may include pedestrians crossing the road away from the zebra crossing or say, maybe a car entering pedestrian area. Since, the moving objects are responsible for such events, the idea is to focus primarily on the foreground and obtain segments of the moving objects. To serve the purpose of foreground extraction, we use ViBe technique as described in \cite{ViBe}. A similar method of foreground extraction has been adopted in \cite{pathak2015anomaly}, but for a very different purpose i.e. for formation of visual words for topic model. In addition to being faster, the advantage of ViBe is that it doesn't let the object in foreground fade away in the background quickly even when the object has stopped moving. This ensures that the objects are still highlighted as foreground in such events, for instance, when a pedestrian halts in the middle of the road for some time or an event where a car is being parked at a no-parking zone. We further apply morphological transformations for denoising. Thus, we segment different blobs of foreground which we use for the purpose as described next.



    


\subsubsection{Learning to Segment Through Motion and Noise}
We follow an approach similar to Pathak {\it et al.} \cite{pathakCVPR17learning} to segment the objects. We use fully convolutional networks for semantic segmentation with default parameters \cite{FCN,teichmann2016multinet}. The foreground obtained from ViBe technique provide us with the pseudo ground truth for training. Moreover, the ViBe's output might not always lead to desired segments and is prone to noise. 
It is evident from extensive experiments in \cite{pathakCVPR17learning} that CNN is able to learn well even from noisy and often incorrect ground truth. Because of its finite capacity, a CNN will not be able to over-fit the noise. Instead, it closely learns the underlying correct segmentation leading to much smoother and visually more correct segmentation compared to the pseudo ground truth. 

\begin{figure}[t]
\begin{minipage}[t]{0.49\linewidth}
    \includegraphics[width=\linewidth]{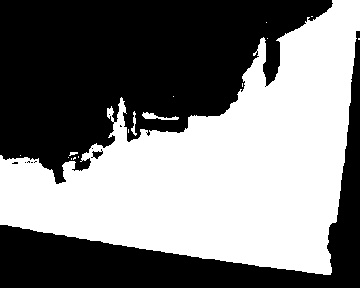}
    \centering
    \small{(a) Traffic Junction Dataset \cite{varadarajan2009topic}}
    \label{static1}
\end{minipage}
\begin{minipage}[t]{0.49\linewidth}
\includegraphics[width=\linewidth]{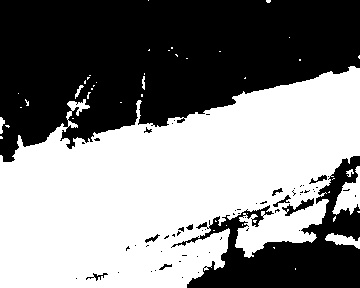}
\centering
\small{(b) Highway Dataset \cite{pathak2015anomaly}}
    \label{static2}

\end{minipage} 
\caption{Representative binary motion mask map for overall foreground motion in the surveillance video datasets.}
    \label{fig:still}
    \vspace{-5mm}
\end{figure}

\subsubsection{Sample Transformation}

Once we segregate foreground from the background using ViBe motion segmentation, we obtain various blobs for each video frame. We find the rectangular connected blob of pixels using contour detection algorithm. 
These connected components represent the moving objects such as pedestrians and cars in video. There might be small groups of pixels as well in the frame which are not likely to be the actual objects of interest.  We use area of those blobs as a parameter to eliminate noise, if any. 

Generally, anomalous events arise due to presence of objects or actions which we usually do not expect. In the Traffic Junction Dataset \cite{varadarajan2009topic} , car entering pedestrian area, jay walking or people crossing the road away from the zebra crossing are among the unusual events as shown in Figure \ref{traffic-dataset}. This motivates our idea to crop the candidate object and place it to other parts of the same frame to make the frame appear abnormal. The object is confined within the rectangle which contains its background as well. If we crop the whole rectangular region, it would look unnatural when we place it on to other parts of the frame. Instead, we apply semantic segmentation \cite{FCN} to the rectangular part using our trained model discussed earlier in Section \RomanNumeralCaps{3} \textit{(A.1} and \textit{A.2)} Thus, we now obtain a precise foreground object after segmentation.

After obtaining the segment, it needs to be placed in other parts of the frame in an unsupervised way. We place objects only in those regions which have observed motion in the entire video. The intuition behind the algorithm is that it is highly unlikely for a car or a person to be present within a building's wall or flying in air. Again, we achieve this by leveraging ViBe motion segmentation. We pre-process the video by keeping track and storing all those pixel indexes which have witnessed motion as guided by ViBe. We achieve this task in real time. As shown in Figure \ref{fig:still}, white pixels represent the regions where there has been foreground object motion at least once and the black pixels represent the regions that never observed any motion. We now have the set of candidate pixels (white pixels in Figure \ref{fig:still}) on or in the vicinity of which the foreground objects shall be placed.

Having cropped the object and placed its segmentation to another region in the frame, the next challenge lies in inpainting that particular missing area in the frame. The objective of region inpainting is to ensure that the synthesized frame appears visually natural. A basic heuristic has been proposed by Bertalmio \textit{et al.} in \cite{bertalmio2001navier} to provide an inpainting algorithm. This technique works well when the region to be inpainted is very small in size. However, as the missing region size is large, so is the case for size of real world objects, it fails to provide a smooth and clear inpainted region. 

\begin{figure}[h!]
\begin{minipage}[t]{0.49\linewidth}
    \includegraphics[width=\linewidth]{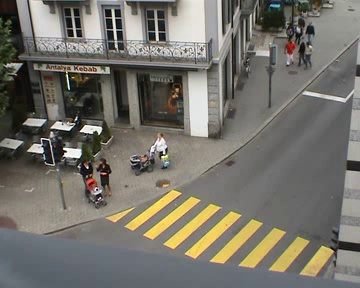}
    \centering
    \label{f1}
\end{minipage}
\begin{minipage}[t]{0.49\linewidth}
\includegraphics[width=\linewidth]{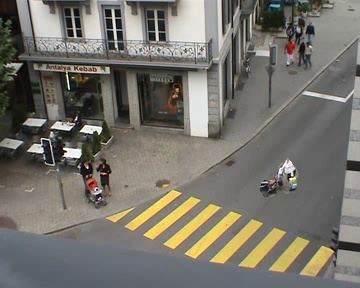}
  
    \label{f2}
\end{minipage} 

\vspace{-0.2cm}

\begin{minipage}[t]{0.49\linewidth}
    \includegraphics[width=\linewidth]{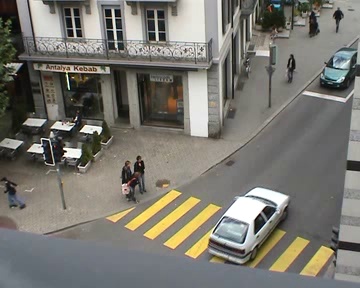}
    \centering
    \label{f1}
\end{minipage}
\begin{minipage}[t]{0.49\linewidth}
    \includegraphics[width=\linewidth]{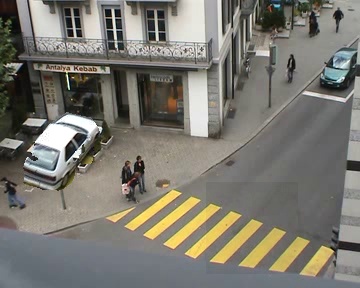}
    \label{f2}
\end{minipage} 

\vspace{-0.2cm}
 
\begin{minipage}[t]{0.49\linewidth}
    \includegraphics[width=\linewidth]{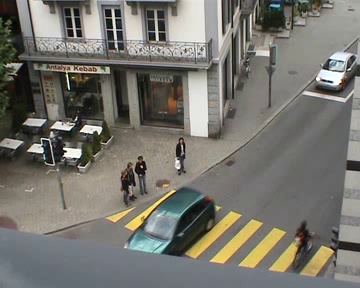}
    \centering
    \label{f1}
\end{minipage}
\begin{minipage}[t]{0.49\linewidth}
    \includegraphics[width=\linewidth]{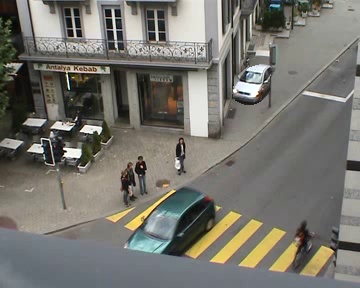}
    \label{f2}
\end{minipage} 

\vspace{-0.2cm}

\begin{minipage}[t]{0.49\linewidth}
    \includegraphics[width=\linewidth]{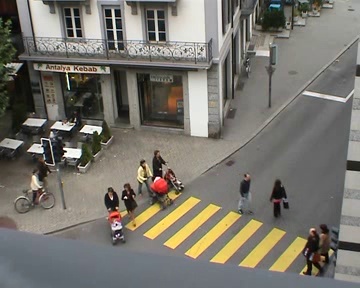}
    \centering
    \label{f1}
\end{minipage}
\begin{minipage}[t]{0.49\linewidth}
\includegraphics[width=\linewidth]{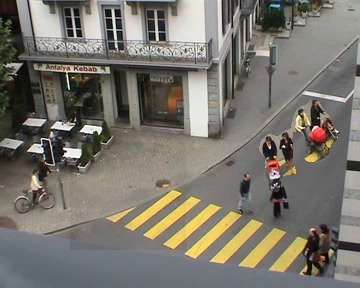}
  
    \label{f2}
\end{minipage} 

  

\vspace{-0.2cm}
\begin{minipage}[t]{0.49\linewidth}
    \includegraphics[width=\linewidth]{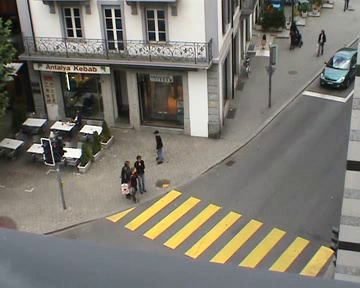}
    \centering
    \label{f1}
\end{minipage}
\begin{minipage}[t]{0.49\linewidth}
\includegraphics[width=\linewidth]{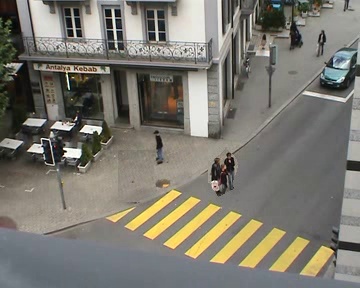}
    \label{f2}
\end{minipage} 
\vspace{-0.2cm}

\begin{minipage}[t]{0.49\linewidth}
    \includegraphics[width=\linewidth]{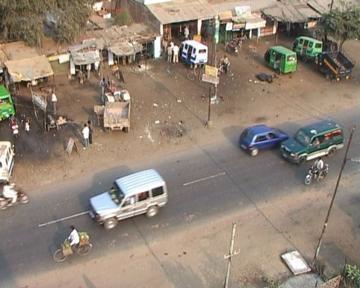}
    \centering
    \label{f1}
\end{minipage}
\begin{minipage}[t]{0.49\linewidth}
\includegraphics[width=\linewidth]{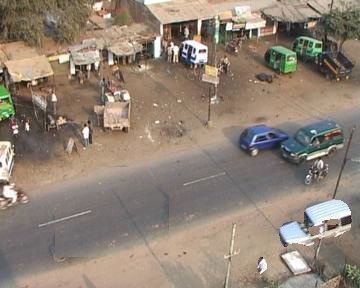}
  
    \label{f2}
\end{minipage} 
\vspace{-0.2cm}

\caption{Sample frames from Traffic Junction and Highway Dataset annotated as \textit{normal} (left) and our \textit{synthesized} abnormal frames (right) from the normal.}
\vspace{-0.2cm}

\label{synthesised}
\end{figure}

We tackle this challenge of region inpainting with a simple approach. First, we exploit the fact that there are plenty of instances when no foreground object is present at a particular region of the frame in the entire video.\\ 
Second, while the surveillance camera is static, the background remains almost unchanged throughout in the video. Our main idea here is to observe the missing rectangular region in previous video frames and choose the frame which has the least or ideally, no movement. We use ViBe for real time pixel level motion information. We cache a constant number of frames arriving earlier than the current frame and observe the ViBe pixels within the missing rectangular region's coordinates in those cached frames. Among these frames, the region having the least motion pixels count indicated by ViBe is chosen as the region that will replace the missing region in the current frame. This approach neatly replaces the missing region compared to the one inpainted using the algorithm in \cite{bertalmio2001navier}. The qualitative results of our proposed algorithm for synthesizing abnormal frames are shown in Figure \ref{synthesised}. 

\subsection{Abnormal Activity Detection}

The sole purpose of generating abnormal frames, was to have a self-supervised learning mechanism to detect abnormal activities. We create our own dataset consisting of abnormal frames and normal frames as training samples. We can synthesize abnormal frames from each normal frame of the video. We train a Bayesian VGG-19 \cite{simonyan2014very} for the two classes.

We observe that while preparing the dataset, synthesis of abnormal frames does not always lead to a sample that actually appears to be abnormal. This is justifiable because the foreground object will be stochastically placed to other regions of a frame. For example, a car might be placed in the road itself, which is actually not abnormal. To overcome this problem, we introduce a Bayesian discriminator into the pipeline. This helps in statistically sampling a refined set of fake samples that can further be used to build a robust classifier.

\begin{figure}[t]
\begin{minipage}[t]{0.49\linewidth}
    \includegraphics[width=\linewidth]{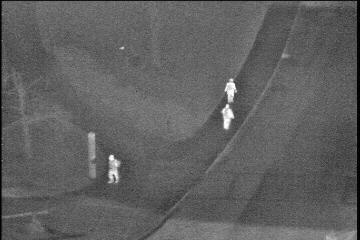}
    \centering
    \label{thermal1}
\end{minipage}
\begin{minipage}[t]{0.49\linewidth}
\includegraphics[width=\linewidth]{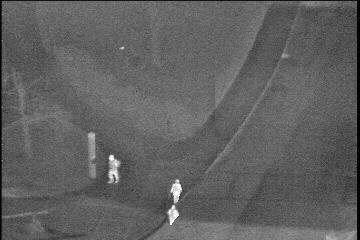}
    \label{thermal1abn}

\end{minipage} 
\vspace{-0.3cm}

\begin{minipage}[t]{0.489\linewidth}
    \includegraphics[width=\linewidth]{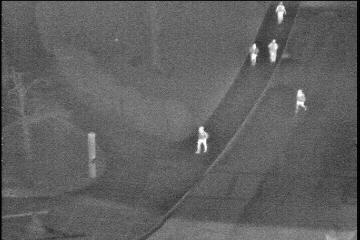}
    \centering
    \label{thermal1}
\end{minipage}
\begin{minipage}[t]{0.499\linewidth}
\includegraphics[width=0.98\linewidth]{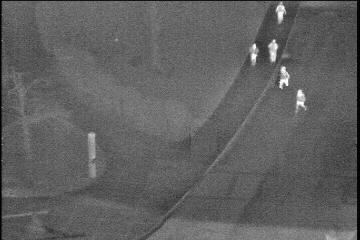}
    \label{thermal1abn}
\end{minipage}

\vspace{-0.2cm}

\caption{Our Synthesis Block technique has another application as a strategy for Data augmentation as well. Sample frames from OSU thermal imagery Dataset~\cite{OTCBVS} (left) and \textit{synthesized} frames (right) by our algorithm.}
\label{Thermalsynthesised}
\vspace{-4mm}
\end{figure}

We interpret dropout as approximate Bayesian stochastic inference \cite{Gal_ARX2015} over our model weights. Moreover, to impose Bernoulli distributions over model weights, we have adopted dropout \cite{srivastava2014dropout}. 
Modeling Bernoulli distributions can be achieved by sampling multiple times (ten times, in our case) with dropping random units during inference. This sampling is possible because it is a Bayesian stochastic inference. These samples are considered as the Monte Carlo samples \cite{kendall2015bayesian} from the posterior distribution over models. We experimented with various dropout ratios and use the following values for the same. For implementing Bayesian CNN, we use dropout ratio of [0.1, 0.1, 0.3, 0.4, 0.4] for each stack of convolutional layers respectively and 0.5 for FC layers. As the number of neurons increases in subsequent layers, we increased the dropout ratio for better generalization. The predicted posterior probability is obtained by applying the softmax function on the network output layer.

We draw ten samples of each synthesized image $I_{i}$ using Monte-Carlo sampling from a distribution (this is predictive posterior distribution for the bayesian CNN). The predictive posterior probability signifies the probability of the sample belonging to the normal class. We calculate the mean $\mu_{i}$ and $\sigma_{i}$ for image $I_{i}$. 
\begin{equation}
  p'_{i} = \mu_{i} - \frac{\sigma_{i}}{2}
  \label{eq8}
\end{equation}

We consider $p'_{i}$ as an upper bound on the predictive posterior probability as criteria to incrementally train the Bayesian CNN and use it to classify abnormal frames. As shown in Figure \ref{fig:size-plot}, we vary the predictive posterior probability $p'_{i}$ and iteratively sample abnormal frames having probability $\le p'_{i}$. Our assumption here is that as there are large number of placements that are abnormal and the real samples are all normal, the model is able to distinguish the abnormal from the normal with some uncertainty. A few examples of the synthesized abnormal samples are provided in Figure \ref{synthesised}. 

\section{Experiments and Results}

We perform experiments on the Traffic Junction Dataset released in \cite{varadarajan2009topic}. It consists of a video of 45 minutes recorded at a frame rate of 25 fps with frame size of $288\times360$. 

For comprehensive evaluation of the proposed algorithm on a real world surveillance video dataset, we perform experiments on the Highway Dataset released in \cite{pathak2015anomaly}. The dataset consists of 6 and a half minute video having frame rate 25 \textit{fps} with frame size of $288\times360$, illustrating the traffic scene of a highway in real world situation.

\begin{figure}[b]
\vspace{-5mm}
\includegraphics[ width=1.0\linewidth]{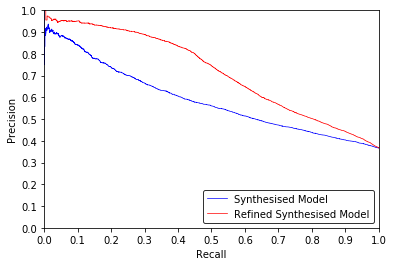}
    \centering
    
    \caption{Precision-Recall curves for anomaly detection in Traffic-Junction Dataset \cite{varadarajan2009topic} using Bayesian CNN pipeline. Anomalous frames were considered as positive and the Normal frames as negative examples.}
    \label{fig:auc-plot}
\end{figure}


\begin{table}[t]
\caption{Results for Anomaly Detection. The reported values are area under the precision-recall curve (AUC)}
\begin{center}
  \centering
\begin{tabular}{ |c|c|c|c|c| }
 \hline
  \textbf{Dataset } & \textbf{Model} & \textbf{AUC} \\ 
  \hline
& Likelihood Model \cite{varadarajan2009topic} & 54.47 \\
 Traffic-Junction~\cite{varadarajan2009topic} & Projection Model \cite{pathak2015anomaly} & 65.15 \\\cline{2-3}
 & \textbf{Our Proposed (non-Bayesian)} & \textbf{74.29}\\
& \textbf{Our Proposed (Bayesian)} & \textbf{ $\mu$: 77.61} , $\sigma$: 0.167 \\[1ex] 
\hline
& Likelihood Model \cite{varadarajan2009topic} & 67.30\\
Highway~\cite{pathak2015anomaly} & Projection Model \cite{pathak2015anomaly} & 81.40 \\\cline{2-3}
& \textbf{Our Proposed (non-Bayesian)} & \textbf{83.59}\\
& \textbf{Our Proposed (Bayesian)} & \textbf{ $\mu$: 68.87} , $\sigma$: 1.2\\
 \hline
\end{tabular}
\end{center}
\label{results}
\vspace{-3mm}
 \end{table}

Additionally, in Figure \ref{Thermalsynthesised} we show that our Synthesis Block has application in data augmentation for thermal imagery as well. OSU Thermal Pedestrian Database~\cite{OTCBVS} is a publicly available benchmark dataset 
which consists of only 284 images. 
Deep learning algorithms, however, don't perform well with such datasets having few data samples. So, our synthesis technique could help in training deep learning algorithms.


Anomalous video frames were separated from the video for testing. From the remaining normal set of frames, 25\% of the normal frames were also included in the test data along with the anomalous ones. For training the CNN, we used our synthesized dataset. Moreover, while adding the normal frames to the synthetic training set, we choose to skip frames by using frame rate of 5 fps instead of 25 fps. This would lower the number of similar frames in the train set, thereby, reducing redundancy.

After having synthesized the abnormal frames, we now have the dataset consisting of two classes: \textit{normal} and \textit{synthesized abnormal}. 
We train a CNN with dropouts to have an approximate Bayesian stochastic inference as described in previous section. We train VGG-19 \cite{simonyan2014very} on this dataset with image size $224\times224$. We use a weighted cross-entropy loss function for training in order explicitly weigh more penalties (five times more, in our case) for false negatives than false positives and SGD as the optimizer. We call this \textit{synthesized model}.

\begin{figure}[t]
\includegraphics[ width=1\linewidth]{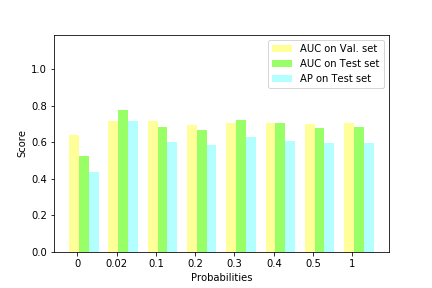}
    \centering
    \caption{Scores of AUC and Average Precision (AP) for Traffic-Junction Dataset \cite{varadarajan2009topic} on incrementally training Bayesian CNN by varying the refined set of samples in the synthesized abnormal class.}
    \label{fig:size-plot}
    \vspace{-5mm}
\end{figure}

To further make our model robust, we obtain a refined dataset by removing the more probable normal fake examples. We apply \textit{softmax} activation function on our trained model to obtain respective classification probabilities (predictive posterior) on the training synthesized abnormal samples. A low probability score for a sample frame to belong to the normal class could indicate a higher likeliness for it being abnormal.  Thus, we refine our synthesized set of examples and iteratively train the Bayesian CNN on this set. Initially, we choose $p'_{i}=0$ and vary it till $p'_{i}=0.5$, thereby, adding more samples to incrementally train the CNN. Note that, at $p'_{i}=1$ in Figure \ref{fig:size-plot} is actually our initially proposed \textit{synthesized model}.

The experimental results on appropriately choosing the clean set of abnormal examples are shown in Figure \ref{fig:size-plot}. We adopt hold-out cross validation technique with validation set consisting of $30\%$ of the dataset. The model with highest AUC performance on the validation set is chosen to report the results by evaluation on the test set. 
Additionally, we evaluate our other models obtained by varying the selected set of examples on the test set too. The results are consistent with the results in validation. Moreover, our proposed algorithm backed with experimental results suggests that actively choosing the training examples further assists in improving the results as shown in Figure \ref{fig:auc-plot}. Furthermore, we experiment with a non-Bayesian setting, keeping our overall pipeline fixed except that we now do not adopt a Bayesian CNN and replace the discriminator with ResNet-50 \cite{he2016deep} representing a non-Bayesian setting. The improved ResNet architecture results in an improvement of the area under the precision-recall curve (AUC) as shown in Table \ref{results}.  As can be seen from the quantitative results, we obtain consistent improvements in terms of AUC. 
The AUC being the standard metric shows improved performance which can be improved by further refining various blocks of the network.

\section{Analysis}

In this section, we provide analysis for our proposed model from aspects such as distribution discrepancy, feature visualization and qualitative analysis.

\subsection{Distribution Discrepancy: proxy $\mathcal{A}$-distance}

According to domain adaptation theory \cite{ben2010theory}, $\mathcal{A}$-distance as a measure of cross domain discrepancy, which, together with the source risk, will bound the target risk. The proxy $\mathcal{A}$-distance is defined as $d_{\mathcal{A}} = 2(1-2\epsilon)$ where $\epsilon$ is the generalization error of a classifier(e.g. kernel SVM \cite{boser1992training}) trained on the binary task of discriminating source and target. We can apply a similar concept to better analyze our results quantitatively. We use our model features and use it to calculate the proxy $\mathcal{A}$-distance i.e. $d_{\mathcal{A}1}$ between two classes: our synthesized abnormal class and the ground truth abnormal class. Moreover, we compare that with the proxy $\mathcal{A}$-distance i.e. $d_{\mathcal{A}2}$ between two classes: our synthesized abnormal class and the normal class. We obtain the values $d_{\mathcal{A}1} = 1.8589$ and $d_{\mathcal{A}2} = 1.9782$. Since, $d_{\mathcal{A}1} < d_{\mathcal{A}2}$. which suggests that our features can reduce the cross-domain gap more effectively. In other words, it gives an indication that features learned for our synthesized abnormal samples would be more closer to the real anomalous video frames.

\begin{figure}[h]

\vspace{-4mm}
  \begin{minipage}[t]{0.45\linewidth}
    \includegraphics[width=\linewidth]{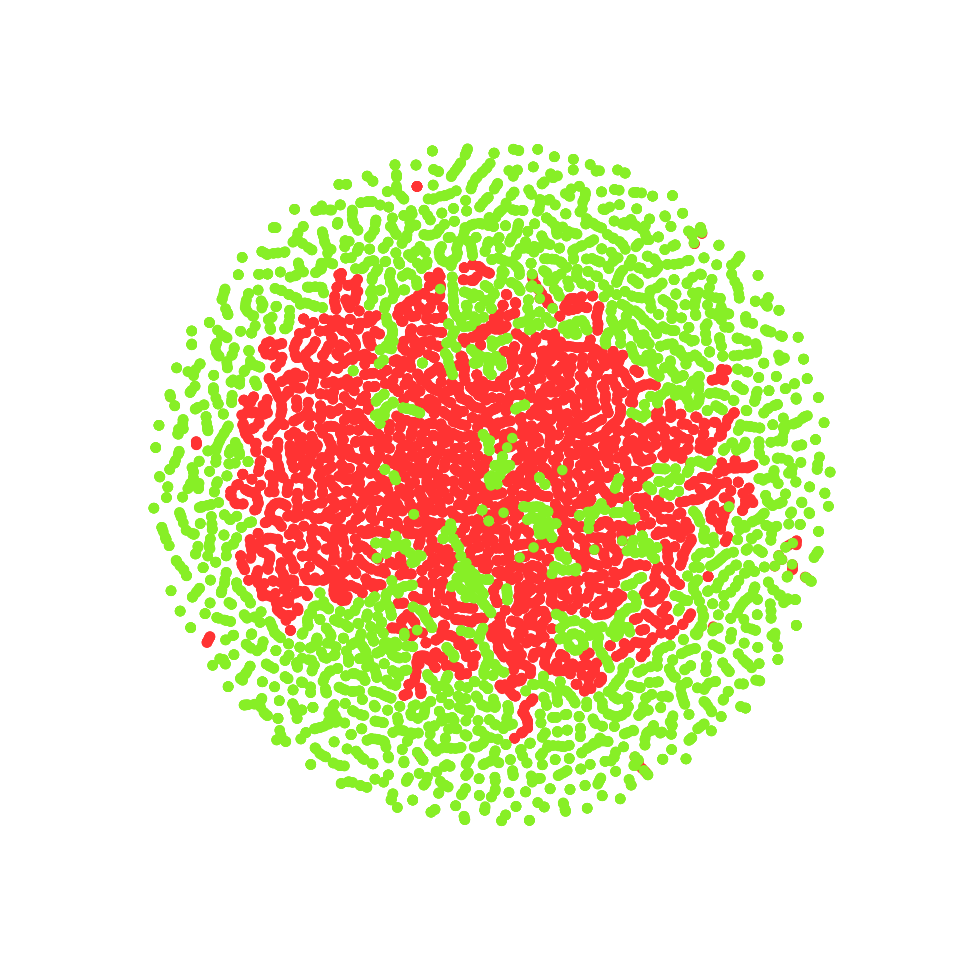}
    \centering
    \small{(a)}
    \label{fig:tsne1}
  \end{minipage}
  \begin{minipage}[t]{0.45\linewidth}
    \includegraphics[width=\linewidth]{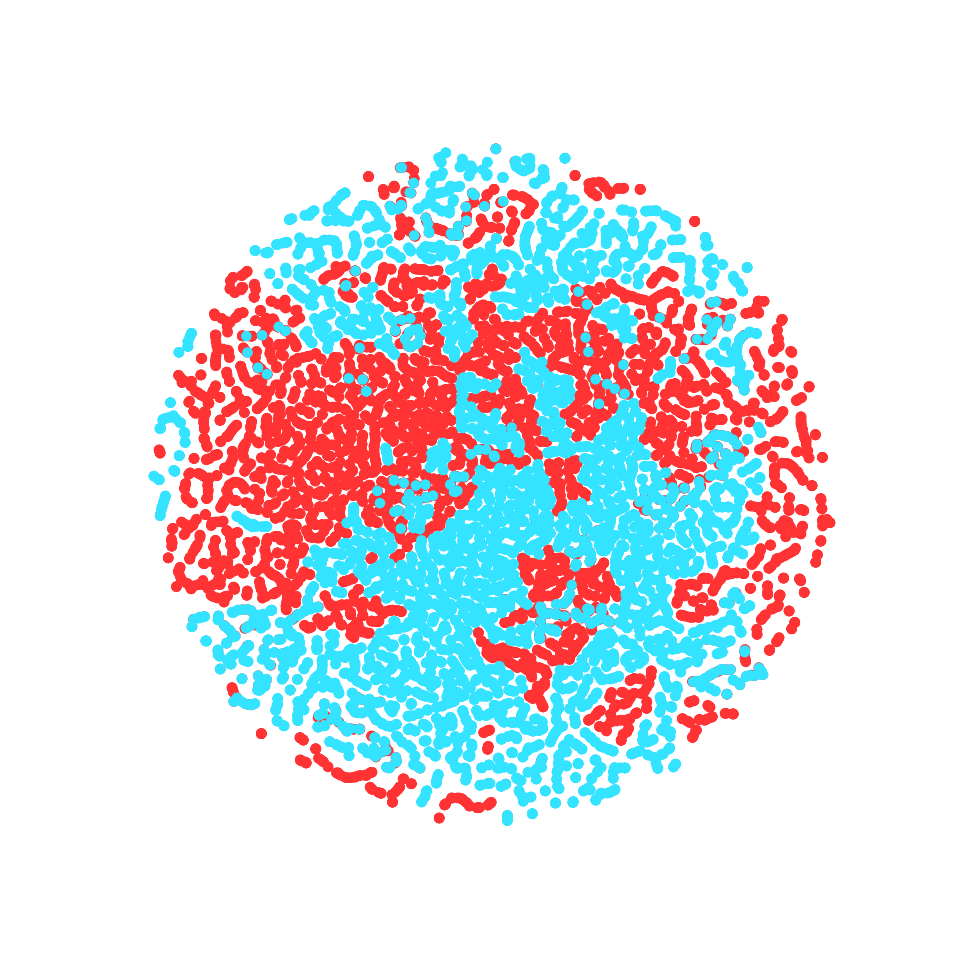}
    \centering
    \small{(b)}
    \label{fig:tsne2}
  \end{minipage}

  \caption{t-SNE visualizations for following distributions: \newline (a) Green points represent \textit{Normal video frames} distribution The centrally clustered red points represent our \textit{Synthesized frames} distribution. \newline (b) Blue points represent \textit{Ground truth abnormal} distribution, while the red points represent our \textit{Synthesized} distribution for Traffic-Junction Dataset~\cite{varadarajan2009topic}.}
\label{fig:tsne}
\end{figure}

\subsection{t-SNE Plot}

t-distributed Stochastic Neighbor Embedding (t-SNE) \cite{maaten2008visualizing} is a technique to visualize high-dimensional data. It converts similarities between data points to joint probabilities and tries to minimize the Kullback-Leibler \cite{Kullback_AMS1951} divergence between the joint probabilities of the low-dimensional embedding and the high-dimensional data. In Figure \ref{fig:tsne}(a), the red points correspond to our synthesized abnormal samples, while the green ones correspond to the normal frames in the video. It is clearly observed that the synthesized samples are mostly clustered towards the centre. 

\begin{figure}[b]
\vspace{-3mm}
  \begin{minipage}[t]{0.3\linewidth}
    \includegraphics[width=\linewidth]{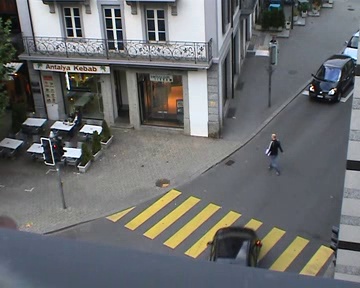}
    \centering
        \small{(a)} 
    \label{fig:1}
  \end{minipage}
  \begin{minipage}[t]{0.3\linewidth}
    \includegraphics[width=\linewidth]{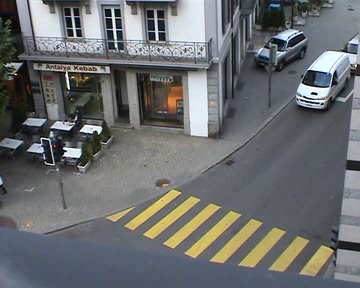}
    \centering
        \small{(b)} 
    \label{fig:2}
  \end{minipage}
  \begin{minipage}[t]{0.3\linewidth}
    \includegraphics[width=\linewidth]{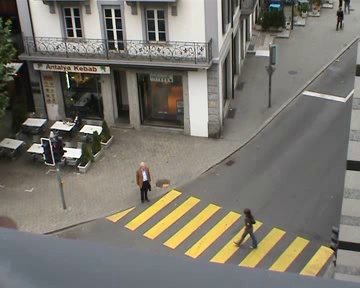}
\centering
\small{(c)} 
    \label{fig:1}
  \end{minipage}
  \caption{Experiments analysis on the Traffic-Junction Dataset \cite{varadarajan2009topic} Figures (a) and (b) have been detected abnormal while (c) has not been detected as abnormal by our model.}
\label{fig:exp_analysis}
\vspace{-5mm}
\end{figure}

In Figure \ref{fig:tsne}(b), the red points correspond to our synthesized abnormal samples, while the blue ones correspond to the ground truth abnormal samples. The red points are spread across the blue points as shown in Figure \ref{fig:tsne}(b). Through the t-SNE \cite{maaten2008visualizing} visualizations in Figure \ref{fig:tsne}, it can be observed that the feature representation of our synthesized abnormal samples makes it much closer to that of the actual ground truth samples distribution as shown in Figure \ref{fig:tsne}(b) and far from the embeddings of the normal video frames since the points are mostly clustered at the centre as shown in Figure \ref{fig:tsne}(a).

\subsection{Qualitative Analysis}
In the 44 minutes long video of the Traffic Junction Dataset\cite{varadarajan2009topic}, there is an instance shown in Figure \ref{fig:exp_analysis} \textit{(a)} where abnormality occurs in the form of a jay walker for a short period of 3 seconds. Our model is able to detect such an activity, which could be tiresome or could have gone undetected by a human operator for such surveillance videos. Also, the frame where a car enters the pedestrian area has also been detected as abnormal as shown in Figure \ref{fig:exp_analysis} \textit{(b)}. However, the model fails to detect the frame in Figure \ref{fig:exp_analysis} \textit{(c)} as abnormal for which ground truth label is abnormal. This is a special case as it features a pedestrian crossing the road at the crossing. The context is that the pedestrian is doing so when the traffic signal is red. It is therefore an abnormal example. However, as this is a very subtle cue even for humans, it is very challenging to detect such cases.

\vspace{-3mm}
\section{Conclusion} \vspace{-1mm}
In this paper, we have presented a method that uses transformation of the normal for generating abnormal samples. These are then validated through a classifier to automatically obtain a set of abnormal examples. This is a novel approach as other approaches such as conditional GANs were also not suited for generating such abnormal examples, but our proposed pipeline is able to generate abnormal examples automatically without supervision. In future, we would be interested in considering spatio-temporal generation of such synthetic examples. We can also further consider other scenarios such as social interactions and imbalanced activity recognition cases where such approaches may be applicable. 
\vspace{-1mm}

{\small
\bibliographystyle{ieee}
\bibliography{egbib}
}



\end{document}